\pdfoutput=1

\documentclass[11pt]{article}

\usepackage[]{acl}

\usepackage{times}
\usepackage{latexsym}

\usepackage[T1]{fontenc}

\usepackage[utf8]{inputenc}

\usepackage{color}
\usepackage{tabularx}
\usepackage{tabu}
\usepackage{booktabs}
\usepackage{multirow}

\usepackage{graphicx} 
\usepackage{float}
\usepackage{subfigure} 
\usepackage{amssymb}

\usepackage{hyperref}
\usepackage{tablefootnote}
\usepackage{xspace}
\usepackage[dvipsnames]{xcolor}

\usepackage{float}
\usepackage{amsmath}
\usepackage{bm}
\usepackage{algorithmicx}
\usepackage{algorithm}
\usepackage{algpseudocode}
\usepackage[most]{tcolorbox}
\usepackage{arydshln}

\usepackage{amssymb}
\usepackage{pifont}
\usepackage{enumitem}

\usepackage{microtype}

\newcommand{\ours}{\textsc{LimRank}\xspace}
\newcommand{\synth}{\textsc{LimRank-Synthesizer}\xspace}

\newcommand{\ie}{\hbox{\emph{i.e.,}}\xspace}

\newcommand{\followir}{\textsc{FollowIR}\xspace}
\newcommand{\bright}{\textsc{Bright}\xspace}
\newcommand{\rankone}{\textsc{Rank1}\xspace}
\newcommand{\reasonrank}{\textsc{ReasonRank}\xspace}
\newcommand{\diver}{\textsc{DIVER}\xspace}

\newcommand{\msmarco}{MS MARCO\xspace}

\newcommand{\github}{\raisebox{-1.5pt}{\includegraphics[height=1.05em]{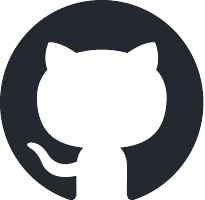}}\xspace}

\title{\ours: Less is More for Reasoning-Intensive Information Reranking}

\author{
Tingyu Song\thanks{~Equal Contributions. Correspondence: Yilun Zhao (\texttt{yilun.zhao@yale.edu}).} \quad 
Yilun Zhao$^*$ \vspace{5pt} \quad Siyue Zhang \quad Chen Zhao \quad Arman Cohan \\
Yale NLP Lab
}

\begin{document}
\maketitle
\begin{abstract}
Existing approaches typically rely on large-scale fine-tuning to adapt LLMs for information reranking tasks, which is computationally expensive. In this work, we demonstrate that modern LLMs can be effectively adapted using only minimal, high-quality supervision. 
To enable this, we design \synth, a reusable and open-source pipeline for generating diverse, challenging, and realistic reranking examples.
Using this synthetic data, we fine-tune our reranker model, \ours.
We evaluate \ours on two challenging benchmarks, \ie \bright for reasoning-intensive retrieval and \followir for instruction-following retrieval. 
Our experiments demonstrate that \ours achieves competitive performance,
while being trained on less than 5\% of the data typically used in prior work.
Further ablation studies demonstrate the effectiveness of \synth and the strong generalization capabilities of \ours across downstream tasks, including
scientific literature search and retrieval-augmented generation for knowledge-intensive problem solving.

\vspace{-5pt}
\begin{center}
\begin{tabular}{rl}
\github & \href{https://github.com/SighingSnow/LimRank}{\path{SighingSnow/LimRank}}\\
\end{tabular}
\end{center} 

\end{abstract}

\section{Introduction}

\begin{figure}[!t] 
    \centering
    \includegraphics[width=0.97\linewidth]{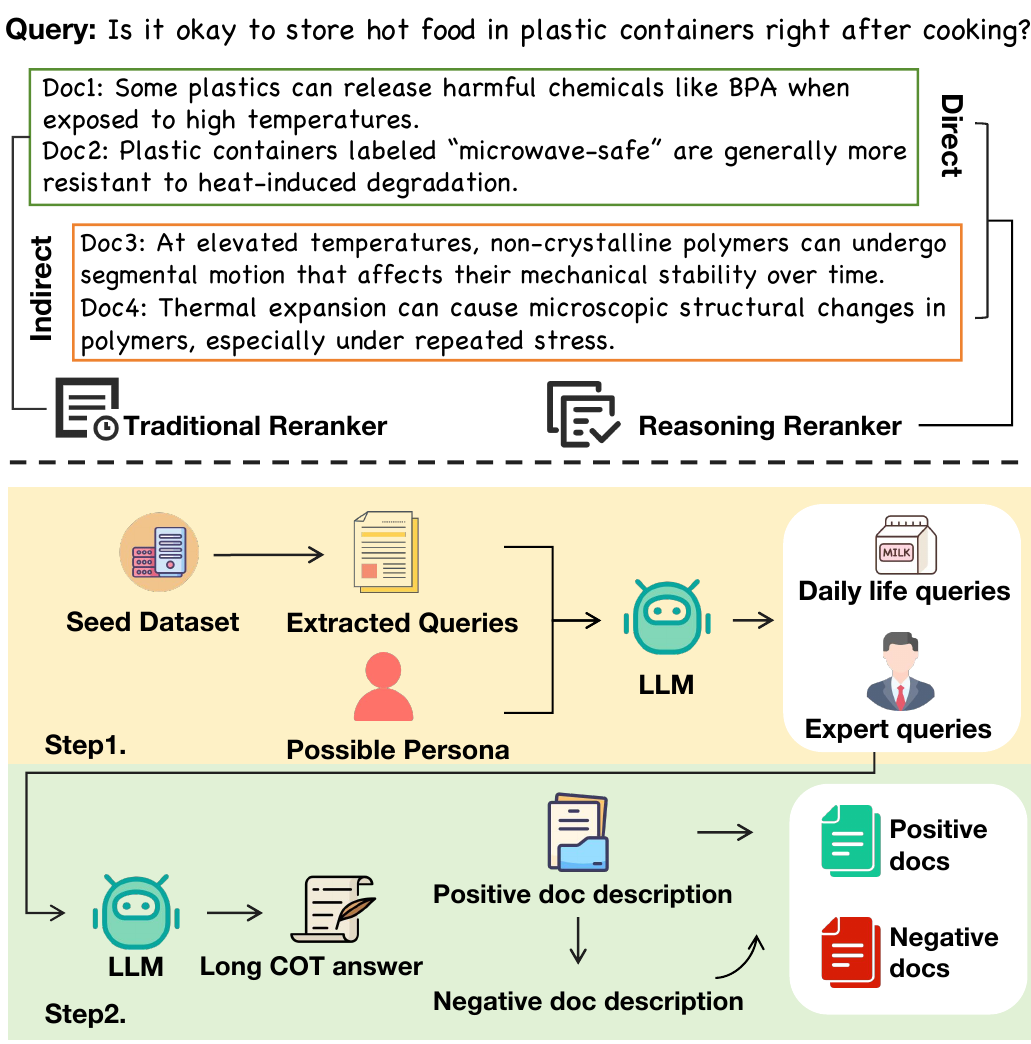}
    \caption{
(Top) An illustration of reasoning-intensive reranking scenarios that demand more than surface-level semantic matching. 
These tasks require multi-step inference, 
contextual reasoning, 
and recognition of implicit relationships between queries and documents.
(Bottom) An overview of \synth, which generates high-quality training data for reranking tasks.
}
    \label{fig:tea}
\end{figure}
Recent studies have increasingly leveraged LLMs for reranking tasks in information retrieval~\cite{peng2025efficiency,  rankr1}. 
While LLMs have shown effectiveness in general-purpose reranking scenarios, emerging research reveals notable limitations when these models are applied to reasoning-intensive retrieval settings~\cite{diff, rank1, reasonir, song-etal-2025-ifir, zhang2025mrmr}. 
These limitations extend beyond performance metrics. They arise from the inherent challenges LLMs encounter when relevance depends on more than surface-level keyword matching or shallow semantic similarity. In such scenarios, effective retrieval requires multi-step inference, contextual reasoning, and the ability to recognize implicit relationships between queries and documents~\cite{bright}.

Inspired by the growing success of reasoning LLMs~\cite{o1, r1}, several recent studies~\cite{rank1, rankr1, yan2025o1embedder}
have begun exploring the training of LLMs that can leverage test-time computation to improve performance in reasoning-intensive retrieval tasks. 
However, contemporary approaches typically rely on large-scale supervised fine-tuning. 
For instance, Rank1~\cite{rank1} is fine-tuned on over nearly 700K examples of DeepSeek-R1's reasoning traces, 
which is expensive in terms of compute and data.

We hypothesize that frontier LLMs already possess considerable latent reasoning capabilities for reranking, and that these capabilities can be activated and steered using a small number of carefully curated, high-quality examples that encourage extended deliberation. This \emph{``less is more''} approach has shown promise in other domains: for example, LIMA~\cite{zhou2023lima} and LIMO~\cite{limo} achieve strong performance in instruction following and complex reasoning with minimal but strategically selected examples, demonstrating that carefully curated demonstrations can effectively steer pretrained models without the need for massive fine-tuning.
To our knowledge, we are the first to investigate this paradigm in reranking tasks.

We introduce \synth, a modular and open-source pipeline for generating high-quality reranking training data through several novel design choices.
\synth is guided by three core principles: domain diversity, alignment with real-world use cases, and difficulty diversity. It generates retrieval queries paired with corresponding positive and negative passages, and employs frontier reasoning models 
(\ie DeepSeek-R1) to produce multi-step reasoning traces. 
We then apply LLM-based filtering to discard low-quality traces. 
Using \synth, we generate a compact yet effective dataset of 20K examples—only 2.85\% of the data used in Rank1.
We fine-tune Qwen2.5-7B on this dataset to produce our reranker model, \ours. 

We evaluate \ours on two challenging reranking tasks: (1) reasoning-intensive retrieval using \bright~\cite{bright}, and (2) instruction-following retrieval using \followir~\cite{followir}.
\ours achieves the nDCG@10 score of 28.0\% on \bright and $p$-MRR score of 1.2 on \followir, representing the best performance among models with 7B-level parameters. 
To better understand the strengths and limitations of \ours, we conduct an in-depth human evaluation, revealing that \ours excels particularly in settings that require multi-hop reasoning, subtle instruction disambiguation, and context-sensitive reranking. 
Additionally, we conduct extensive ablation studies on \synth\ and show that each component of our guidelines is essential. We also train different model variants using synthetic IR datasets of the same size (\ie 20K). Models trained with our data consistently outperform those trained with other synthetic IR datasets used by \rankone, Promptriever, and ReasonIR.

To assess \ours's practical utility, we further deploy and evaluate it in two real-world-inspired tasks: (1) Scientific Literature Search on the LitSearch dataset~\cite{litsearch}, and (2) Retrieval-Augmented Generation (RAG) on the GPQA benchmark~\cite{rein2024gpqa}. 
Compared to the previous state-of-the-art, Rank1-7B, \ours demonstrates strong generalization, achieving 30.3\% accuracy on GPQA (vs. 28.3\%) and a competitive 60.1\% Recall@5 on LitSearch (vs. 60.8\%), demonstrating its effectiveness as a plug-in reranker in real-world systems.

\section{Related Works}\label{sec:related-work}

Reasoning-intensive Information Retrieval~\cite{bright} has gained increasing attention due to its practical relevance in real-world scenarios.  
Some works, such as \rankone~\cite{rank1} and ReasonIR~\cite{reasonir}, address this challenge by fine-tuning models on large-scale, reasoning-focused retrieval datasets to enhance the performance of rerankers and retrievers, respectively. 
Recently, other works~\cite{r2r, abdallah2025dear, liu2025reasonrank} have proposed complex training strategies using synthetic data. However, the large data volume required by these methods leads to high training costs, and the training strategies themselves introduces extra training cost. 
Some other works~\cite{searchr1, r1searcher, rankr1} employs reinforcement learning to equip retrievers and rerankers with reasoning capabilities. While effective, these methods often entail significant computational, data, and engineering overhead.  
Recent studies~\cite{zhou2023lima, limo} suggest that modern LLMs possess substantial latent reasoning abilities across a range of tasks, which can be efficiently activated and enhanced using small amounts of task-specific training data. 
To support this, various data selection methods~\cite{less, lessvl} have been proposed and shown to be effective. 
To our best knowledge, our work is the first to apply such \emph{``less is more''} exploration to reranking tasks. 
\begin{table*}[!ht]
\centering
\scalebox{0.8}{
\begin{tabular}{p{1.7cm}p{17cm}}
\toprule
\textbf{Element} & \textbf{~~Data Generation Guideline} \\
\midrule
\textbf{Query} &
\begin{tabular}{p{17cm}}
\emph{\textbf{Domain Diversity:}} Queries should not limited to everyday contexts but also span specialized domains such as finance, law, and healthcare.\\[4pt]
\emph{\textbf{Alignment with Daily:}} Life Queries vary in complexity to mirror real-world use cases. Simple queries reflect straightforward needs, while complex ones mimic intricate scenarios.\\[4pt]
\emph{\textbf{Difficulty Diversity:}} The dataset should contain simple queries with challenging ones. Hard queries incorporate patterns like instruction-following, multi-step reasoning, and so on.
\end{tabular}
\\
\midrule
\textbf{Reasoning Chain} &
\begin{tabular}{p{17cm}}
\emph{\textbf{Well-Structured Reasoning:}} Reasoning chains should be organized logically to enhance inference efficiency. Clear step-by-step processes help the model navigate solutions systematically during search operations.\\[4pt]
\emph{\textbf{Human Verification:}} Reasoning chain undergoes human review to validate accuracy, coherence, and alignment with the query’s intent. This ensures robust, error-resistant logic.\\[4pt]
\emph{\textbf{Adaptive Analysis:}} For simple queries, reasoning focuses on explicit connections between the query and passage. For complex queries, chains prioritize identifying implicit relationships, resolving ambiguities, and addressing layered contextual demands
\end{tabular}
\\
\midrule
\textbf{Passage} &
\begin{tabular}{p{17cm}}
\emph{\textbf{Hard Negatives:}} Hard negative passages with subtle alterations can help train the model in discerning fine-grained distinctions.\\[4pt]
\emph{\textbf{Complexity Diversity:}} Passages vary in structure and content difficulty, including lengthy texts, analytical data (e.g., statistical reports), or domain-specific jargon.
\end{tabular}
\\
\bottomrule
\end{tabular}
}
\caption{Data generation guideline. We encourage the query to be diverse. Consequently, the reasoning chain will be more diverse. We believe this can activate LLM potential more easily. }
\label{table: Data Generation Guidelines}
\end{table*}

\section{\synth}

\subsection{Data Curation Guidelines}
\label{data-curation-guideline}
\synth is designed to unlock the LLM’s latent potential by minimizing the number of training examples while maximizing their informativeness and depth of reasoning.
We establish a set of guidelines to govern the construction of training data, as detailed in~\autoref{table: Data Generation Guidelines}.

\subsection{Reasoning-intensive Data Generation}
Commonly used training datasets for IR tasks (\ie \msmarco) are limited in complexity. 
The queries in these datasets are typically straightforward, 
which can constrain model performance on reasoning-intensive retrieval tasks. 
Recently various data synthesis methods~\cite{chen2023few, shen2025hopweaver} are proposed in retrieval and fact checking domains. 
Some works~\cite{almeida2024exploring, sinha2025don} focus on constructing the dataset with higher quality or with better efficiency.  
Recent studies such as \reasonrank~\cite{liu2025reasonrank} and \diver~\cite{long2025diver} have demonstrated strong results on reasoning-intensive tasks using specialized synthetic data. 
A limitation of these methods is their dependency on large-scale data aggregation from varied sources. This work, therefore, focuses on creating a straightforward and easy-to-implement pipeline. 
We propose \synth, a bottom-up data synthesis pipeline designed to generate high-quality training data. 
As illustrated in~\autoref{fig:tea}, \synth uses \msmarco as the seed dataset and generates enhanced data based on its simple queries, guided by our designed guideline discussed earlier. We detail the query generation and passage generation processes as follows:

\begin{table*}[htbp]
\centering

\resizebox{\textwidth}{!}{
\begin{tabular}{lrrrrrrrrrrrrr}
\toprule[.1em]
 & \multicolumn{7}{c}{StackExchange} & \multicolumn{2}{c}{Coding} & \multicolumn{3}{c}{Theroem-based} & \multirow{2}{*}{Avg} \\
  \cmidrule(lr){2-8} 
 \cmidrule(lr){9-10}
 \cmidrule(lr){11-13}
 & Bio. & Earth. & Econ. & Psy. & Rob. & Stack. & Sus. & Leet. & Pony & Aops & TheoQ. & TheoT. & \\

\midrule
BM25s & 18.2 & 27.9 & 16.4 & 13.4 & 10.9 & 16.3 & 16.1 & \textbf{24.7} & 4.3 & \underline{6.5} & 7.3 & 2.1 & 13.7 \\
Monot5-3B & 37.9 & \underline{45.7} & \underline{24.1} & 34.3 & 17.6 & 24.1 & 25.1 & \underline{18.2} & 21.5 & 4.9 & 19.8 & 21.3 & 24.5 \\
RankZephyr-7B & 21.9 & 23.7 & 14.4 & 10.3 & 7.6 & 13.7 & 16.6 & 6.5 & \underline{24.7} & \textbf{6.8} & 2.0 & 7.3 & 13.0 \\
RankGPT & 33.8 & 34.2 & 16.7 & 27.0 & \textbf{23.3} & \textbf{27.7} & 11.1 & 15.6 & 3.4 & 1.2 & 8.6 & 0.2 & 17.0 \\
RankLLaMA-7B & 40.9 & 44.0 & 23.6 & \underline{35.6} & 21.1 & 23.5 & 25.6 & 16.5 & 23.3 & 3.7 & 14.2 & 14.7 & 23.9 \\
Rank-R1 & 26.8 & 24.8 & 17.9 & 22.1 & 17.4 & 10.3 & 21.1 & 4.4 & 15.6 & 3.3 & 10.4 & 5.9 & 15.0 \\
Qwen2.5-7B & \underline{50.7} & 38.5 & 21.0 & 30.0 & 16.0 & 21.1 & 22.9 & 18.0 & 15.3 & 3.7 & 17.4 & 16.3 & 22.6 \\
\rankone-7B & 48.8 & 36.7 & 20.8 & 35.0 & 22.0 & 18.7 & \textbf{36.2} & 12.7 & \textbf{31.2} & 6.3 & \textbf{23.7} & \textbf{37.8} & \underline{27.5} \\
\ours & \textbf{52.5} & \textbf{49.8} & \textbf{25.4} & \textbf{43.1} & \underline{22.4} & \underline{25.7} & \underline{27.9} & 17.5 & 20.2 & 4.7 & \underline{20.0} & \underline{27.3} & \textbf{28.0} \\
\bottomrule[.1em]
\end{tabular}
}
\caption{nDCG@10 results on \bright. All methods rerank the top-100 documents retrieved by BM25. Results for BM25, RankZephyr-7B, RankGPT4, and RankR1 are copied from Rank-R1. }
\label{table-bright-results}
\end{table*}

\paragraph{Query Generation.}
Recognizing that individuals from different backgrounds focus on different aspects of a topic, we use a variety of personas to ensure query diversity. We first randomly sample several personas from PersonaHub~\cite{persona} and use a query sampled from the seed dataset to prompt the LLM to generate a persona that suits the context.
Given the same query and a specific persona, we then prompt an LLM (\ie GPT-4o) to generate two types of augmented queries: one situated in a daily-life context and the other in an expert-domain context. 
For daily-life scenarios, we encourage the LLM to rewrite queries such that they involve more complex or nuanced real-world situations, 
that may challenge everyday human reasoning. 
For expert-domain scenarios, we focus on questions requiring domain-specific knowledge. These include queries seeking evidence, challenging a claim, or reasoning about complex professional issues. 
All generated queries are derived from the original \msmarco queries, using carefully designed prompts detailed in Appendix~\ref{query-prompt}.

\paragraph{Positive and Negative Passage Generation.}
We construct passages that are both directly and indirectly related to the query.
To achieve this, we employ the CoT prompting technique to guide the LLM in describing the intermediate materials needed to solve the problem. 
The final positive passage is then generated based on these descriptions.
As prior work~\cite{e5} has shown, hard negative passages are crucial for effective training. To create such negatives, we provide the LLM with the query and the associated positive descriptions, and then prompt it to generate corresponding hard negative descriptions. And these negative descriptions are used in the same way as positive's to generate negative passages. 
The prompts used for both positive and negative generation are included in Appendix~\ref{passage-prompt}.

\subsection{Data Filtering}
Following \rankone, we ask DeepSeek-R1 to judge the relevance between query and the passage. 
If the judgment is the same as we generated previously, then we collect this dataset into our final dataset and filter the rest of them. Finally, we collected 10,282 (query, passage) pairs. 
\section{Experiments}
This section presents our main results and ablation studies. We provide details on the experimental setup, including evaluation setting, baseline systems, and implementation specifics, in Appendix~\ref{app:exp-setup}.

\subsection{Main Results} 
\paragraph{BRIGHT Results.} 
We present results on \bright in ~\autoref{table-bright-results}. 
Compared to \rankone, 
which adopts a similar approach to \ours for training a pointwise reranker, 
\ours surpasses its performance and achieves state-of-the-art results on this task. 
Although the improvement is modest, it is noteworthy that our model is trained with significantly less data than \rankone (20K vs. 6 million examples), 
This not only reduces training cost but also partially alleviates the efficiency concerns associated with \rankone. 
Moreover, when compared to other models, \ours shows a clear performance advantage. 
Listwise models like RankGPT4 and RankerZephyr are not explicitly trained for reasoning-intensive scenarios, which may account for their lower performance. 
The setwise reranker Rank-R1, trained via RL, 
also falls behind \ours, despite setwise rerankers generally having structural advantages over pointwise models. 
This suggests that either the reward function or the training data used in Rank-R1 may require more careful design and curation. 

\paragraph{\followir Results.} 
Instruction-following has garnered increasing attention in the IR community. 
As shown in ~\autoref{compress-analysis-results}, \ours achieves state-of-the-art performance on \followir. 
Notably, \ours outperforms \rankone while using substantially less training data, which may be attributed to the increased diversity and complexity of the synthesized training data. 
This supports the conclusion that a smaller amount of high-quality data can more effectively activate the model’s instruction-following capabilities.
However, the performance gap between \ours and \followir-7B remains relatively small. 
This implies that test-time scaling alone may not be sufficient for enhancing instruction-following performance, highlighting the importance of carefully designed training data even for large models.

\subsection{Error Analysis}
\label{error-analysis}

\ours fall short in some cases, we conclude them as follows:  
(1) \textbf{Queries requiring directly relevant passages.} When the query clearly implies a need for directly relevant content, \ours typically gives similar scores to both direct and indirect relevant passages. 
(2) \textbf{Queries with very few relevant passages.} In cases where only one or two golden passages exist, \ours exhibits uncertainty during the relevance judgment process and is more likely to fail. 
These observations suggest a potential direction for future work: developing an adaptive reranker capable of distinguishing query intent and assigning relevance scores with greater contextual awareness. 
And we provide a detailed case study in Appendix~\ref{apdx-case-study}.

\begin{table}[!t]
\centering
\small
\setlength{\tabcolsep}{3pt}
\resizebox{0.48\textwidth}{!}{
\begin{tabular}{lrrrr }
\toprule[.1em]

\multirow{2}{*}{Model} & \multicolumn{2}{c}{\followir} & LitSearch & GPQA \\ 
& Score & $p$-MRR & Recall@5 & Acc \\
\midrule
Qwen2.5-7B & 11.6 & -0.8 & 49.5 & 26.3 \\
RankLLaMA-7B & 15.5 & 0.0 & 59.0 & 28.3 \\
FollowIR-7B & 15.6 & 0.3 & 57.5 & \textbf{30.3} \\
\rankone-7B & 16.8 & 0.3 & \textbf{60.8} & 28.3 \\ 
\textbf{\ours} & \textbf{17.5} & \textbf{1.2} & 60.1 & \textbf{30.3} \\ 
\bottomrule[.1em]
\end{tabular}
}

\caption{Average performance on relevant tasks. Detailed results can be found in Appendix~\ref{app:exp-setup}. 
}
\label{compress-analysis-results}
\end{table}

\subsection{Experiments on Real-world Tasks}
\label{downstream-analysis}

To assess the practicality of \ours in real-world applications, we evaluate it on two downstream tasks.
The results are shown in ~\autoref{compress-analysis-results}.
LitSearch~\cite{litsearch} is a retrieval dataset containing natural queries in the scientific literature domain. For the \textit{specific} question type in LitSearch—where no more than five papers are relevant—\ours achieves a Recall@5 of 60.1\%, which is comparable to the 60.8\% achieved by \rankone, indicating only a negligible performance gap.
GPQA~\cite{rein2024gpqa} is a challenging benchmark covering the domains of biology, physics, and chemistry. 
To assess real-world usefulness, we evaluate \ours on this dataset within a RAG setup. \ours achieves state-of-the-art performance with an accuracy of 30.3\%, outperforming \rankone’s 28.3\%. 
Overall, \ours demonstrates comparable performance across downstream tasks 
within less data involved in training. 
Detailed experimental settings and analysis are provided in Appendix~\ref{downstream-exp}.

\subsection{Analysis of Synthetic Training Data} 
\label{ablation-study} 
To validate our methodology, we assess both the individual components of our guidelines and the overall quality of our synthetic pipeline. 

\begin{table}[!t]
\centering
\small
\setlength{\tabcolsep}{1.5pt}
\resizebox{0.49\textwidth}{!}{
\begin{tabular}{lcccc}
\toprule
\textbf{} & \textbf{BRIGHT} & \textbf{FollowIR} & \textbf{LitSearch} & \textbf{GPQA} \\
\midrule
\textbf{\ours (Full Set)} & $\mathbf{28.0}$ & $\mathbf{1.19}$ & $\mathbf{60.1}$ & $\mathbf{30.3}$ \\
\quad Daily-life Queries & $27.6$ & $0.39$ & $59.6$ & $26.8$ \\
\quad Expert Domain Queries & $26.7$ & $0.85$ & $59.1$ & $26.8$ \\
\quad Short Reasoning Trace & $27.0$ & $0.72$ & $59.9$ & $27.3$ \\
\quad Long Reasoning Trace & $26.7$ & $1.17$ & $59.3$ & $27.3$ \\
\bottomrule
\end{tabular}
}
\caption{Performance comparison across different configurations of our multi-level guideline. \ours refers to the full version. }

\label{tab: ablation-component}
\end{table}

\paragraph{Dataset Design Components.}
From ~\autoref{tab: ablation-component}, we can find that each component of our guidelines is essential. Simple queries from \rankone are insufficient, while daily life queries greatly boost reasoning and RAG performance. 
Adding long reasoning traces improves both instruction following and RAG, highlighting the importance of query-adaptive reasoning.

\paragraph{Training Dataset Size. }As shown in~\autoref{compress-analysis-results}, 
\ours achieves strong performance on both reranking and real-world application tasks, despite being trained on far less data. 
This raises the question of whether the scaling law still applies in this context. 
Although some prior works~\cite{limo} demonstrate that even small datasets can yield strong results, we observe that performance plateaus when the data size becomes too limited. 

\begin{table}[!t]
\centering
\small
\setlength{\tabcolsep}{3pt}
\resizebox{0.48\textwidth}{!}{
\begin{tabular}{lcccc}
\toprule
\textbf{} & \textbf{BRIGHT} & \textbf{FollowIR} & \textbf{LitSearch} & \textbf{GPQA} \\
\midrule
Promptriever & $25.5$ & $0.96$ & $59.8$ & $28.8$ \\
ReasonIR & $25.6$ & $0.21$ & $\mathbf{60.4}$ & $26.8$ \\
\textbf{\ours}  & $\mathbf{28.0}$ & $\mathbf{1.19}$ & $60.1$ & $\mathbf{30.3}$ \\
\bottomrule
\end{tabular}
}
\caption{Performance comparison between different synthesis datasets. 
}

\label{tab: ablation-dataset-compare}
\end{table}
 
\paragraph{Effectiveness of \synth. }
To further examine the effectiveness of our synthesized data, we trained several reranker variants using different synthetic IR datasets of the same size (20K examples), including \rankone, Promptriever, ReasonIR, and \ours.
As shown in \autoref{tab: ablation-dataset-compare}, the variant trained with LimRank consistently outperforms those trained on the other synthetic datasets, demonstrating its effectiveness for training rerankers in low-resource settings.

\section{Conclusion}

We propose \synth, a data synthesis pipeline that generates high-quality IR training data.
Trained with far fewer but higher-quality examples, \ours achieves state-of-the-art results on the reasoning and instruction-following benchmarks, and demonstrates strong performance on downstream tasks. 
Overall, \ours provides empirical support for the \emph{``Less is More Hypothesis''} in the IR field, offering a cost-effective alternative to data-intensive approaches.  

\section*{Limitations}
In this section, we highlight two limitations of our work, each of which offers promising directions for future research. 
First, our data selection method is both naive and costly. As described earlier, we rely on human verification during the data generation process and apply basic filtering techniques. While effective, this approach is resource-intensive and may not scale efficiently. Developing automated verification strategies or more sophisticated filtering methods could significantly improve efficiency and reproducibility in future work.
Second, we apply our synthesized training data only to pointwise rerankers. Although our method shows strong performance in this setting, it remains unexplored how well the data would transfer to other architectures, such as listwise or setwise rerankers. Extending our pipeline to support these paradigms may further enhance generalization and flexibility.
These limitations suggest several directions for future work. 


\bibliography{anthology,custom, llm}

\appendix

\section{Prompts}
\label{prompts}

\subsection{Prompt for Query Expansion}
\label{query-prompt}
We provide the prompts used for query generation as follows. 
First, we incorporate personas from PersonaHub~\cite{persona} to personalize the queries. 
Then, as described previously, we expand each query into two types: one set reflecting real-life scenarios and the other targeting expert domains. 
The prompt used for persona integration is shown in~\autoref{fig-prompt-persona}. And we random sample personas from PersonaHub each time. 
We use~\autoref{fig-prompt-daily} for real-life scenario expansion and~\autoref{fig-prompt-expert} for expert domain expansion.

\begin{figure*}[htbp]
\centering
\footnotesize
\fbox{%
    \begin{minipage}{0.9\textwidth}
    Based on the query, please generate a persona similar to the example personas below. \\ 
    
    Query: [FILL\_QUERY\_HERE] \\ 

    Here are some example personas for reference: 
[FILL\_EXAMPLES\_HERE] \\ 

Please provide only the personalized information relevant to the query, without any additional content. 
    
    \end{minipage}%
}
\caption{Prompt for query expansion in daily life scenario. }
\label{fig-prompt-persona}
\end{figure*}

\begin{figure*}[htbp]
\centering
\footnotesize
\fbox{%
    \begin{minipage}{0.9\textwidth}
        Given a query and a persona, please concrete the query by incorporating the persona. 
Query: [FILL\_QUERY\_HERE] \\ 
Persona: [FILL\_PERSONA\_HERE] \\ 

Here are some requirements for the concretized query: \\    
1. The new query should not be a direct combination of the query and the persona.  \\
2. It should be a newly formed query influenced by the persona.  \\ 
3. Don't asking too many questions. Please focues only 1-2 questions.  \\ 

And you should provide a context or scenario related to the new query. \\ 
Here are some requirements for the scenario: \\ 
1. The context and scenario should be narrated in first person or narrate some facts. \\
2. The context or scenario can be the background or the situation which lead to the new query.  \\ 
3. The context or scenario should be within a main line and in a logical order. \\ 
4. The context or scenario should have some controversial points leading to the new query. But you should not include "controversial" and similar words in the context or scenario. \\

You should return with in the following format: \\ 
```json\{ \\ 
    "query" : "the new query", \\ 
    "scenario" : "the scenario or context related to the new query" \\
\}``` \\ 
    \end{minipage}%
}
\caption{Prompt for query expansion in daily life scenario. }
\label{fig-prompt-daily}
\end{figure*}

\begin{figure*}[htbp]
\centering
\footnotesize
\fbox{%
    \begin{minipage}{0.9\textwidth}
        Given a query, please make a new query. \\ 

Query: [FILL\_QUERY\_HERE] \\ 

Here are some requirements for the new query: \\ 
1. The new query should be in the same field as the original query. But the new query should be a different topic. For example the original query is in the history field, the new query should be in the history field but a different topic.  \\ 
2. The new query should be a more professional query that can include some professional terms or jargons.  \\ 
3. The new query should be a more complex query. But it should mainly focus on 1-2 points.  But 1 point is better.  \\ 
4. The new query should be expanded like one's thinking process. It should show one's exploration or thinking in an expert-domain and final lead to the new query.  \\ 
5. The new query can include statistical number or formula, and some research papers or news.  \\

So what you should do is: \\ 
1. In-breathe the query to another topic but in the same field. And the query should be a more professional query. The in-breathing query should focus mainly on 1-2 small points.  \\ 
2. Expand the new query like one's thinking or exploration process.  The person can be influenced by some papers, politics, or some events. And someone maybe confused by the conflict or contradiction between their recognition and the facts.  These should be included in the newly generated query. \\ 

For example, \\ 
1. As far as I understand, overfitting in machine learning happens when a model learns the training data too well, including its noise, and thus performs poorly on unseen data. Regularization techniques like dropout or L2 loss are supposed to prevent overfitting by penalizing complexity. \\ 
But in practice, large models trained on huge datasets—like GPT-style models—seem to avoid overfitting even when they're massively overparameterized. Why don’t they overfit despite having more parameters than training samples? \\ 
If it's due to the scale of data, then shouldn't smaller models trained on large data also generalize as well? And if it’s about optimization dynamics, why do simple regularizers help at all in small-scale settings? \\ 
2. The fall of the Western Roman Empire in 476 AD is commonly seen as the end of ancient Rome and the beginning of the Middle Ages. The deposition of Romulus Augustulus by Odoacer is often cited as the decisive event. \\ 
But Roman institutions, language, and even laws continued to exist in various forms across Europe for centuries—especially in the Byzantine Empire, which still called itself Roman. If Rome "fell," why did so much of its structure persist? \\ 
If the fall of Rome was a definitive collapse, why didn’t it lead to a more immediate or total cultural break? And if it was more of a transformation, should we even think of it as a “fall” at all? \\

Please only provide the new query, without any additional content.\\ 
    \end{minipage}%
}
\caption{Prompt for query expansion in daily life scenario. }
\label{fig-prompt-expert}
\end{figure*}

\subsection{Prompt for Passage Generation}
\label{passage-prompt}
As mentioned earlier, we first use~\autoref{fig-prompt-problem-sovle} to prompt the LLM (\ie GPT-4) to solve the query using a detailed chain-of-thought (CoT) reasoning process. 
We then extract relevant passage descriptions using~\autoref{fig-prompt-desc-extraction}. 
To ensure that the negative passages are sufficiently challenging, we provide the LLM with the query and the positive passage description, and use~\autoref{fig-prompt-neg-pos-desc} to generate hard negative descriptions. 
Finally, we generate the corresponding positive and negative passages using~\autoref{fig-prompt-psg-gen} for each given descriptions.

\begin{figure*}[htbp]
\centering
\footnotesize
\fbox{%
    \begin{minipage}{0.9\textwidth}

Given a query, please analyze and answer this question, providing the reasoning process for each step. Respond point by point. The response should include the materials needed for the analysis. Please think step by step. \\ 

Query: [FILL\_QUERY\_HERE] \\ 

Please return as detailed a response as possible, including the reasoning process for each step. 

...
    \end{minipage}%
}
\caption{Prompt for problem solving. }
\label{fig-prompt-problem-sovle}
\end{figure*}

\begin{figure*}[htbp]
\centering
\footnotesize
\fbox{%
    \begin{minipage}{0.9\textwidth}
        Given a passage, please extract the materials described in the passage.  \\ 
Here is the passage: 
[FILL\_PASSAGE\_HERE] \\ 

You need to extract the material description. The number of the material description should be in range 3-7.  \\ 
What you extracted need to cover different aspects of the origin passage. And please return in the following format:  \\ 
1. [extracted material description 1] \\ 
2. [extracted material description 2] \\ 
...
    \end{minipage}%
}
\caption{Prompt for materical description extraction.  }
\label{fig-prompt-desc-extraction}
\end{figure*}

\begin{figure*}[htbp]
\centering
\footnotesize
\fbox{%
    \begin{minipage}{0.9\textwidth}
        You are assigned a task in the text information retrieval field. 
Given a query, positive passage descriptions, you need to generate negative passage descriptions. \\ 

Query: [FILL\_QUERY\_HERE] \\ 
Positive passage descriptions: [POSITIVE\_PASSAGE\_DESCRIPTIONS\_HERE] \\ 

The negative passage descriptions should follow the following guidelines: \\ 
1. The negative passage description can be partially related to the query's topic. \\ 
2. The negative passage description can be used to generate hard negative passages.  \\ 
3. You should consider the given positive passage descriptions. The negative passage description should be different from the positive passage descriptions. \\

You can return 1-5 negative passage descriptions. And you should return in a list format as follows: \\ 
1. [negative passage description 1] \\ 
2. [negative passage description 2] \\
...
    \end{minipage}%
}
\caption{Prompt for query expansion in daily life scenario. }
\label{fig-prompt-neg-pos-desc}
\end{figure*}

\begin{figure*}[htbp]
\centering
\footnotesize
\fbox{%
    \begin{minipage}{0.9\textwidth}
        Given a material description, please generate a passage that satisfies the material description. \\ 

Here is the material description:
[FILL\_MATERIAL\_DESCRIPTION] \\ 

The generated passage should satisfy the following requirements: \\ 
- Satisfying the material description.  \\ 
- Include the number or analysis if the material description requiring analysis or calculation.  \\ 
- Include the policy or principle if the material description requiring policy or principle. \\
- Include the legal cases or laws if the material description requiring legal cases or laws. \\ 
- Include the examples or evidence if the material description requiring examples or evidence. \\ 

Anyway, please generate a passage that satisfies the material description. \\ 
Please only provide the passage, without any additional content. 
    \end{minipage}%
}
\caption{Prompt for passage generation.  }
\label{fig-prompt-psg-gen}
\end{figure*}

\subsection{Prompt for R1}
\label{r1-prompt}
The prompt we used for DeepSeek-R1 to get the reasoning chain and the final relevance judgement is the same as Rank1. It is shown in ~\autoref{fig-prompt-r1}.

\begin{figure*}[htbp]
\centering
\footnotesize
\fbox{%
    \begin{minipage}{0.9\textwidth}
Determine if the following passage is relevant to the query. Answer only with 'true' or 'false'. \\
Query: [FILL\_QUERY\_HERE] \\ 
Passage: [FILL\_PASSAGE\_HERE] \\
    \end{minipage}%
}
\caption{Prompt for Deepseek-R1 judgement.  }
\label{fig-prompt-r1}
\end{figure*}

\section{Experiment Settings and Details}
\label{app:exp-setup}

\subsection{Main Experiments Settings}
\paragraph{\ours Settings. } 
We adopt Qwen2.5-7B from the Qwen2.5~\cite{qwen25} series as our backbone model. 
The training data consists of two components: 14,000 examples from \msmarco used in Rank1~\cite{rank1}, and 6,000 examples generated by \synth. 
We ensure that positive and negative passages are balanced within the training dataset. 
Additional training details are provided in Appendix~\ref{training-settings}.

\paragraph{Datasets. } 
We evaluate \ours on two key abilities required for reranking: reasoning and instruction following. 
For reasoning, we benchmark performance across all subtasks of \bright. we retrieve the top-100 documents using BM25 and compute nDCG@10 on the reranked results. 
To assess instruction-following ability, we evaluate on \followir. Unlike the original \followir setup, we retrieve only the top-100 documents using BM25 instead of the top-1000. 
Except the number of documents to rerank, we follow the same evaluation settings as in \followir. 
We mainly calculate top-5 documents' scores in nDCG, MAP and $p$-MRR. We provide detailed results in Appendix~\ref{followir-exp}. 
We further provide an analysis on the downstream task LitSearch~\cite{litsearch} and GPQA~\cite{rein2024gpqa} to assess real-world applicability. For LitSearch, we mainly benchmark \ours on the ``Specific'' type questions in LitSearch and calculate Recall@5. For the GPQA RAG experiment, we use the data store constructed in ReasonIR~\cite{reasonir} as the corpus. And we choose the diomand subset for test.  
We ask the reranker to rerank the top-100 documents retrieved from BM25. We use the Qwen2.5-7B as the reader and set topk to be three. We provide a further analysis of the RAG experiment in Appendix~\ref{downstream-exp}. 

\paragraph{Baselines. } 
We use BM25~\cite{bm25} as the initial retriever to obtain the top-100 candidate documents. 
For a fair comparison, we primarily evaluate \ours against models with similar parameter sizes. 
We include listwise rerankers such as RankerZephyr-7B~\cite{rankzephyr} and RankGPT4~\cite{rankgpt}, as well as the setwise reranker Rank-R1~\cite{rankr1}. 
Additionally, we include MonoT5~\cite{monot5}, a widely adopted pointwise reranking model, as a classic baseline.

\subsection{Training and Hyperparameter Details}
\label{training-settings}

We primarily follow the experimental settings from \rankone. 
All experiments are conducted on machines with 2×80GB H100 GPUs. 
Models are trained for 5 epochs, with each training run taking less than 2 hours. 
We use LLaMA-Factory for LoRA fine-tuning on all parameters, with a rank of 32 and alpha set to 64. 
The learning rate is set to 6e-5, and the batch size is 128.

\subsection{Ablation Experiments}
\label{app: ablation-exps} 

\paragraph{Training Dataset Size} We describe the settings of our data scaling experiments, with results shown in~\autoref{table-bright-scaling-results}. 
For the 2K setup, we use 1,000 samples from the \rankone training data and 1,000 samples from \synth. 
For the 10K setup, we use 6,000 samples from \rankone and 4,000 from \synth. 
For the 20K setup, we use 14,000 samples from \rankone and 6,000 from \synth. 
All models in these experiments are trained under the same settings.

\begin{table*}[htbp]
\centering

\resizebox{0.9\textwidth}{!}{
\begin{tabular}{lrrrrrrrrrrrrr}
\toprule[.1em]
 & \multicolumn{7}{c}{StackExchange} & \multicolumn{2}{c}{Coding} & \multicolumn{3}{c}{Theroem-based} & \multirow{2}{*}{Avg} \\
  \cmidrule(lr){2-8} 
 \cmidrule(lr){9-10}
 \cmidrule(lr){11-13}
 & Bio. & Earth. & Econ. & Psy. & Rob. & Stack. & Sus. & Leet. & Pony & Aops & TheoQ. & TheoT. & \\
\midrule
\ours-2k & 47.8 & 43.0 & 24.2 & 35.1 & 19.8 & 23.0 & 26.0 & \textbf{17.8} & 16.6 & 3.9 & 17.9 & 24.8 & 25.0 \\
\ours-10k & 50.5 & 45.3 & 24.7 & 37.6 & 20.5 & \textbf{26.3} & \textbf{28.2} & 17.5 & 16.6 & 4.3 & 19.6 & \textbf{28.0} & 26.6 \\
\ours-20k & \textbf{52.5} & \textbf{49.8} & \textbf{25.4} & \textbf{43.1} & \textbf{22.4} & 25.7 & 27.9 & 17.5 & \textbf{20.2} & \textbf{4.7} & \textbf{20.0} & 27.3 & \textbf{28.0} \\
\bottomrule[.1em]
\end{tabular}
}
\caption{Scaling data quantity influences. nDCG@10 results on \bright.  }
\label{table-bright-scaling-results}
\end{table*}

\paragraph{Dataset Design Components. } We conduct extra ablation experiments in rebuttal stage to validate that each point in the guideline is necessary.  We choose variables like, simple(hard) queries, daily(expert) queries, short(long) reasoning trace. All experiments are conducted with the same 20k training data size. We find that: (1) Simple queries from RANK1 is not enough. (2) Daily life queries can bring huge improvement in reasoning-intensive tasks and RAG scenarios.  But it may fails on instruction-following instructions. (3) Long reasoning trace can benefit instruction following abilities and performance in the RAG performance. (4) This emphasizes the importance of including reasoning traces with query-adaptive lengths in the training data.  And we will provide a detailed analysis in the revised manuscripts. 

\paragraph{Effectiveness of Our Dataset } To further examine the effectiveness of LimRank data, we trained several reranker variants using different synthetic IR datasets of the same size, including \rankone, Promptriever, ReasonIR, and \ours. All models were trained under identical settings to ensure a fair comparison. As shown in the results, the variant trained with \ours consistently outperforms the other synthetic datasets, 
demonstrating its effectiveness compared to other methods in training rerankers in a low-resource setting.  

\paragraph{Importance of Reasoning Traces. }
Under the 20K-example setting, we evaluate both the presence and absence of reasoning traces across \ours, ReasonIR, and \rankone.
We find that:
(1) Training \rankone with reasoning traces leads to improved performance.
(2) LimRank outperforms both ReasonIR and \rankone in settings with and without reasoning traces. 

\subsection{FollowIR Results}
\label{followir-exp} 
We provide the details of \followir results in ~\autoref{table-followir-full-results}. 
\begin{table*}[htbp]
\centering

\resizebox{0.9\textwidth}{!}{
\begin{tabular}{lrrrrrrrr}
\toprule[.1em]
\multirow{2}{*}{Model} & \multicolumn{2}{c}{Robust04} & \multicolumn{2}{c}{News21} & \multicolumn{2}{c}{Core17} & \multicolumn{2}{c}{Avg} \\
 \cmidrule(lr){2-3}
 \cmidrule(lr){4-5}
 \cmidrule(lr){6-7}
 \cmidrule(lr){8-9}
 & MAP & $p$-MRR & nDCG & $p$-MRR & MAP & $p$-MRR & Score & $p$-MRR \\
\midrule
Qwen2.5-7B & 7.7 & 0.4 & 24.2 & -0.9 & 2.8 & -1.9 & 11.6 & -0.8 \\
MonoT5-3B & 10.6 & 0.2 & 30.0 & 0.0 & 4.6 & -0.1 & 15.1 & 0.0 \\
RankLLaMA-7B & 9.8 & 0.0 & 32.4 & 0.0 & 4.3 & 0.0 & 15.5 & 0.0 \\
FollowIR-7B & 9.7 & -0.0 & 32.8 & 0.3 & 4.1 & 0.5 & 15.6 & 0.3 \\
Rank1-7B & 11.4 & 1.6 & 34.0 & -0.3 & 4.9 & -0.3 & 16.8 & 0.3 \\
\ours & \textbf{11.8} & \textbf{1.7} & \textbf{35.7} & \textbf{1.0} & \textbf{4.9} & \textbf{0.8} & \textbf{17.5} & \textbf{1.2} \\
\bottomrule[.1em]
\end{tabular}
}
\caption{Total results for \followir. The score is measured by MAP@5, nDCG@5 and $p$-MRR on top-100 documents retrieved by BM25. }
\label{table-followir-full-results}
\end{table*}

\subsection{Downstream Task Details}
\label{downstream-exp}

\paragraph{LitSearch Experiment} 
The LitSearch dataset includes two types of questions: ``broaden'' and ``specificity''. 
In this work, we focus on the ``specificity'' questions, which require retrieving the top five most suitable documents. 
This setting imposes a higher standard of precision.

\paragraph{Retrieval-Augmented Generation (RAG) Experiment}
For the RAG experiments, we use diamond subset of GPQA as the question-answering benchmark. 
We use the filtered datastore version from ReasonIR as the retrieval corpus. 
We rerank the top-20 documents and select the top-3 to serve as the input to the generation model.

\section{Examples}
\subsection{\synth Examples}

We provide extra examples of \synth. The query examples are shown in ~\autoref{apdx-table-query-examples}. And the passage examples are provided in ~\autoref{apdx-table-passage-examples}. Provided passages are corresponding to the first query in ~\autoref{apdx-table-query-examples}.

\begin{table*}[!t]
\centering
\small

\scalebox{0.8}{
\begin{tabular}{rlp{3.5cm}p{3.5cm}p{7.5cm}}
\toprule[.1em]
ID & Type & Original Query & Persona & Query \\
\midrule

1 & Daily Query
  & what did komodo dragons do
  & A herpetologist specializing in the behavior and ecology of Komodo dragons, dedicated to understanding their role in their natural habitat and their interactions with other species and the environment. 
  & As a herpetologist focusing on Komodo dragons, I've spent years observing how these fascinating creatures navigate their natural environment. While conducting fieldwork on Komodo Island, I noticed that some researchers argue that Komodo dragons primarily rely on their size and strength to dominate their territory, while others suggest that their sophisticated hunting strategies and interactions with other predators contribute more significantly. This discrepancy in understanding has led me to question the specific behaviors that shape their interactions with both prey and potential competitors in their ecosystem. \\

2 & Daily Query
  & what is a good glucose number
  & A diabetes educator specializing in glucose management, dedicated to helping individuals understand optimal blood sugar levels and their impact on overall health. 
  & What are the optimal blood sugar targets I should aim for to effectively manage glucose levels and avoid potential health issues? \\

3 & Expert Query  
  & what is driverquery.exe
  & A computer systems analyst with a focus on Windows operating system utilities, particularly interested in exploring and educating others about built-in tools like driverquery.exe and how they can be used to optimize system performance and troubleshoot hardware issues. 
  & As far as I am aware, Windows Management Instrumentation (WMI) is a crucial infrastructure for managing data and operations on Windows-based operating systems, providing a standardized way to access system information and resources. It is widely used for system administration, monitoring, and automation tasks. However, WMI also poses security concerns due to its capabilities, such as remote access and execution. Given the increasing sophistication of cybersecurity threats, how do organizations effectively balance the need for WMI usage in system management and automation with the potential security risks it poses? Are there specific security frameworks or best practices in place to mitigate these risks while maintaining system efficiency? Additionally, how do recent developments in cybersecurity research address the vulnerabilities associated with WMI, and what role do statistics on WMI-related security incidents play in shaping these strategies? \\

4 & Expert Query
  & how much does it cost to build a bbq island
  & A DIY enthusiast and outdoor living specialist who is passionate about designing and constructing customized outdoor spaces, with a particular interest in building functional and aesthetically pleasing BBQ islands. They are focused on understanding the costs, materials, and design trends involved in creating the perfect outdoor cooking area. 
  & When designing an outdoor kitchen space, especially a high-end BBQ island, the choice of materials can significantly influence both the functionality and longevity of the structure. While traditional materials like concrete and stainless steel are popular, there is an emerging interest in more sustainable and innovative materials such as recycled composite decking or Corian. However, these materials often come with a different set of maintenance requirements and environmental impacts. How do these alternative materials compare to traditional ones in terms of thermal resistance and durability, particularly when exposed to extreme weather conditions? Furthermore, considering the environmental impact, is the initial higher cost of these eco-friendly materials offset by their durability and reduced maintenance over time? Are there any recent studies or data supporting these claims, possibly influencing trends in the industry towards sustainable building practices? \\

\bottomrule[.1em]
\end{tabular}
}

\caption{Query examples. The examples are generated by \synth.}
\label{apdx-table-query-examples}
\end{table*}

\begin{table*}[!t]
\centering
\small

\scalebox{0.7}{
\begin{tabular}{rlp{3.5cm}p{10cm}}
\toprule[.1em]
ID & Type & Materical Description & Passage \\
\midrule

1 & Direct Positive Document 
  & Information from wildlife documentaries detailing Komodo dragons' hunting and feeding behaviors, including their ambush tactics and diet. 
  & Komodo dragons, the apex predators of their island habitats, exhibit fascinating hunting and feeding behaviors that have been meticulously documented in wildlife documentaries. These formidable reptiles employ ambush tactics, using their stealth and patience to great effect. Often lying in wait along animal trails or near water sources, a Komodo dragon remains motionless, blending seamlessly into the environment with its mottled skin until an unsuspecting prey comes within striking distance. Once the prey is close enough, the dragon launches a swift and powerful attack, using its muscular legs to propel itself forward and its strong, serrated teeth to deliver a fatal bite. The diet of a Komodo dragon is diverse, reflecting its opportunistic nature. They primarily feed on large mammals, such as deer and wild boar, but they are also known to consume smaller prey like birds, snakes, and even carrion. Documentaries often highlight their remarkable ability to consume up to 80\% of their body weight in a single meal, showcasing their dominance as the largest living species of lizard. Despite their size and power, Komodo dragons are not entirely reliant on brute force; their saliva contains a mix of toxic bacteria and proteins that can incapacitate prey, ensuring a successful hunt even if the initial attack does not immediately kill. Such documentaries provide compelling evidence of the Komodo dragon's adaptability and survival strategies, offering an in-depth view of their role in the ecosystem as both predators and scavengers. Through expert commentary and vivid footage, viewers gain a deeper understanding of these ancient reptiles, whose hunting prowess and dietary habits have earned them a fearsome reputation in the natural world.\\

2 & Indirect positive(offering background) 
  & Scientific studies describing the Komodo dragon's unique saliva, which contains venomous proteins that help subdue prey. 
  & Scientific studies have delved into the fascinating biology of the Komodo dragon, revealing that its saliva is not just ordinary; it is laced with venomous proteins that play a critical role in its predatory success. This unique composition of the Komodo dragon's saliva is a subject of intrigue and has been analyzed to uncover the specific proteins responsible for its venomous properties. Research has identified several enzymes, including phospholipase A2, which contribute to the anticoagulant effects and cause rapid blood loss in prey. Additionally, these proteins are known to lower blood pressure, induce shock, and inhibit clotting, making it easier for the Komodo dragon to subdue its prey. Studies have shown that the venom's potency is comparable to that of venomous snakes, providing a significant advantage in the wild. This discovery has challenged previous assumptions that the Komodo dragon's hunting prowess was solely due to bacterial infection from its bite, highlighting the complexity and efficiency of its biological weaponry. \\

3 & Negative Document
  & Studies on the dietary habits of other large reptiles, like crocodiles, without specific reference to Komodo dragons or their unique hunting strategies. 
  & In examining the dietary habits of large reptiles, particularly crocodiles, we find fascinating insights into their feeding strategies and ecological roles. Crocodiles, being apex predators, exhibit a diet that is both diverse and opportunistic. Studies have shown that their diet primarily consists of fish, but they are also known to consume a wide range of prey including birds, mammals, and occasionally other reptiles. For example, research conducted in the Everglades reveals that American crocodiles (Crocodylus acutus) can adapt their diet based on availability, often preying on crustaceans and small mammals during specific seasons when these prey are more abundant. Furthermore, Nile crocodiles (Crocodylus niloticus) in Africa have been documented engaging in cooperative hunting strategies, which involve herding fish into shallow waters to facilitate easier capture. This behavior highlights their adaptability and intelligence in securing food resources. Unlike Komodo dragons, which employ a unique hunting strategy involving venom and ambush, crocodiles rely more on their stealth, powerful jaws, and aquatic prowess to capture prey. This ability to exploit a variety of food sources underscores their role as crucial components of their ecosystems, maintaining the balance by controlling the population of various species. \\

\bottomrule[.1em]
\end{tabular}
}

\caption{Passage examples. The examples are generated by \synth.}
\label{apdx-table-passage-examples}
\end{table*}

\subsection{Error Analysis and Case Study}
\label{apdx-case-study}
We compare the output from \ours and \rankone. As shown in ~\autoref{fig-case-study}, \ours is able to identify indirect relevance between the query and the passage, whereas \rankone marks the same passage as irrelevant.  
This may be attributed to the fact that our training data includes examples of indirect relevance, rather than relying solely on golden passages from traditional datasets. 
In addition, We provide examples in~\autoref{apdx-case-study-1} and ~\autoref{apdx-case-study-2} to demonstrate \ours's capability to distinguish fine-grained instruction variations. 
These cases collectively highlight the reasoning and instruction-following abilities exhibited by \ours.

\begin{figure}[!h]
\centering
\footnotesize
\begin{tcolorbox}[
    width=0.47\textwidth,
    colback=gray!10, colframe=gray!50,
    boxrule=0.8pt,        
    arc=2mm,              
    left=2pt, right=2pt, top=6pt, bottom=6pt 
]
\textbf{[\ours Output]:} \\ 
... But since the query is about making a fake signal, the passage does support the resolution by explaining the existing technology that does that. So, \textbf{even though it's not a direct answer, it's foundational and relevant.} Therefore, the answer should be true.

\noindent\rule{\textwidth}{0.4pt}

\textbf{[\rankone Output]:} \\ 
... Even though both topics relate to the auditory system, the passage doesn't provide information on how to create fake signals. It's more about the medical device implantation rather than the technical or experimental methods to produce and deliver such signals. \textbf{Therefore, the passage doesn't address the query's core concern about generating fake auditory signals, making it irrelevant. }

\end{tcolorbox}
\caption{Output comparison between \ours and \rankone. }
\label{fig-case-study}
\end{figure}

\begin{figure*}[!t]
\centering
\footnotesize
\begin{tcolorbox}[
    width=0.9\textwidth,
    colback=gray!10, colframe=gray!50,
    boxrule=0.8pt,        
    arc=2mm,              
    left=2pt, right=2pt, top=6pt, bottom=6pt 
]
\textbf{[Question]}: \\ 
The question is about how to represent GDP by industry as a network graph, where sectors are connected by flows of money or goods, and seeks research or visualizations that illustrate inter-industry links, sources, drains, and economic loops. 

\textbf{[Passage Summary]}: \\ 
The passage is about the extensive research on input–output models, including their extensions to non-linear relationships, inter-industry clustering, and macro-level economic analysis.

\textbf{[Output]}: \\
Okay, let's tackle this query. The user wants to know if the given passage is relevant to their question about viewing GDP as a network graph. The query mentions using industries like real estate and motor vehicles as examples of how sectors might be connected through transactions, creating a network.\\Looking at the passage, it discusses the input-output model, which is a framework used to analyze economic transactions between industries. The passage talks about extending the model to non-linear relationships, the Hawkins-Simon condition, disaggregation into industry clusters, and empirical studies on coefficient identification. It also mentions extending the Leontief system to general equilibrium and decomposition methods.\\Now, the user's main interest is in creating a network graph of GDP industries, showing connections between sectors. \textbf{The passage doesn't explicitly mention network graphs or visualization}. \textbf{However}, the input-output model is foundational for understanding how industries interact, which is exactly what the user is asking about. The passage's discussion of disaggregating industries and empirical work on coefficients relates to building a detailed network model. Even though the passage doesn't use the term \"network graph,\" the underlying principles of the input-output model are relevant because they provide the structure and data needed to create such a visualization. The mention of general equilibrium and decomposition methods might also support analyzing the sources and drains in the network. Therefore, \textbf{even if the passage doesn't directly address network graphs}, it gives \textbf{foundational knowledge about the economic relationships} that would be necessary for creating one. So, the answer should be true. 

\end{tcolorbox}
\caption{Outputs from \ours. The queries and the passages are given in summary format for clarity. The example shows that \ours can distingush the passage with indirect correlation. }
\label{apdx-case-study-1}
\end{figure*}

\begin{figure*}[!t]
\centering
\footnotesize
\begin{tcolorbox}[
    width=0.9\textwidth,
    colback=gray!10, colframe=gray!50,
    boxrule=0.8pt,        
    arc=2mm,              
    left=2pt, right=2pt, top=6pt, bottom=6pt 
]
\textbf{[Question]}: \\ 
The question is about how airlines claiming to offer carbon-neutral flights explain and demonstrate these claims, specifically regarding their use of alternative fuels, transparency, and accounting methods to ensure that emissions are not simply shifted elsewhere. \\ 
\textbf{[Passage]}: \\
The passage is about the Qantas SAF Coalition Program, which supports the development of a domestic sustainable aviation fuel (SAF) industry in Australia by encouraging corporate contributions to offset the higher cost of SAF and reduce business travel emissions.
\textbf{[Output]}: \\
Okay, let's tackle this. The user is asking whether the provided passage is relevant to the query about how airlines plan to make flights carbon-neutral. The query specifically mentions CNN's article from 2022 about Boeing's ecoDemonstrator jet and the airlines' vows to offer carbon-neutral flights. First, the passage talks about the SAF Coalition Program, which is about supporting the development of sustainable aviation fuel (SAF) in Australia. It mentions Qantas as part of the coalition, where corporations contribute to the cost of SAF, helping to make it more affordable. The program's goal is to reduce emissions by using SAF, which is a key method for carbon neutrality in aviation. Now, the query is asking how airlines explain making flights carbon-neutral. The passage directly discusses a program that involves using SAF to reduce emissions, which is a method mentioned in the query. Even though the passage doesn't explicitly say how airlines communicate their plans, it does provide information on a specific approach (SAF) that airlines might use. Since the query is about how they plan to achieve carbon neutrality, the mention of SAF as a method is relevant. The passage also talks about corporate contributions and making SAF affordable, which relates to the financial aspects of implementing such plans. However, the passage doesn't address transparency or accounting methods, which are part of the user's concern. But the question is whether the passage is relevant, not whether it answers all parts. The presence of SAF as a method is directly related to the query's focus on how airlines plan to make flights carbon-neutral. Therefore, the passage is relevant because it provides information on a specific strategy (SAF) that airlines might use, even if it doesn't detail the communication aspects.

\end{tcolorbox}
\caption{Outputs from \ours. The queries and the passages are given in summary format for clarity. The example shows that \ours can also process the passage with direct correlation.  }
\label{apdx-case-study-2}
\end{figure*}

\end{document}